\documentclass[11pt]{article}
\usepackage{acl}
\usepackage{times}
\usepackage{latexsym}
\usepackage[T1]{fontenc}
\usepackage[utf8]{inputenc}
\usepackage{microtype}
\usepackage{graphicx}
\usepackage{booktabs}
\usepackage{amsmath}
\usepackage{url}

\title{From Code-Centric to Concept-Centric: \\Teaching NLP with LLM-Assisted ``Vibe Coding''}

\author{Hend Al-Khalifa \\
  College of Computer and Information Sciences \\
  King Saud University, Riyadh, Saudi Arabia \\
  \texttt{hend@ksu.edu.sa} \\}

\begin{document}
\maketitle
\begin{abstract}
The rapid advancement of Large Language Models (LLMs) presents both challenges and opportunities for Natural Language Processing (NLP) education. This paper introduces ``Vibe Coding,'' a pedagogical approach that leverages LLMs as coding assistants while maintaining focus on conceptual understanding and critical thinking. We describe the implementation of this approach in a senior-level undergraduate NLP course, where students completed seven labs using LLMs for code generation while being assessed primarily on conceptual understanding through critical reflection questions. Analysis of end-of-course feedback from 19 students reveals high satisfaction (mean scores 4.4-4.6/5.0) across engagement, conceptual learning, and assessment fairness. Students particularly valued the reduced cognitive load from debugging, enabling deeper focus on NLP concepts. However, challenges emerged around time constraints, LLM output verification, and the need for clearer task specifications. Our findings suggest that when properly structured with mandatory prompt logging and reflection-based assessment, LLM-assisted learning can shift focus from syntactic fluency to conceptual mastery, preparing students for an AI-augmented professional landscape.
\end{abstract}

\section{Introduction}

The integration of Large Language Models (LLMs) into software development has fundamentally altered the professional landscape, with tools like GitHub Copilot, ChatGPT, and Claude becoming ubiquitous in industry practice. This shift poses a critical question for computer science education: How do we prepare students for a future where AI-assisted coding is the norm, while still ensuring they develop deep conceptual understanding?

This tension is particularly acute in Natural Language Processing (NLP) education, where students must grasp complex linguistic and computational concepts while implementing increasingly sophisticated algorithms. Traditional lab-based instruction often sees students spending substantial time debugging syntax errors and wrestling with implementation details, leaving limited cognitive resources for understanding the underlying NLP principles.

We introduce \textbf{``Vibe Coding''} as a pedagogical framework that embraces LLM assistance while maintaining rigorous conceptual assessment. Rather than treating LLMs as a threat to learning integrity, we position them as tools that can redirect student attention from low-level implementation concerns to high-level conceptual understanding, if properly scaffolded and assessed.

Our approach features three key components: (1) \textbf{Sanctioned LLM use} for code generation during labs, (2) \textbf{Mandatory prompt logging} to encourage strategic LLM interaction, and (3) \textbf{Reflection-based assessment} that evaluates conceptual understanding rather than code quality. This paper presents the first systematic evaluation of this approach in an undergraduate NLP course.

This study investigates: (1) How does structured LLM-assisted coding (Vibe Coding) affect student engagement and conceptual learning in NLP labs? (2) What are students' perceptions of assessment fairness when grading shifts from code quality to conceptual reflection? (3) What challenges and frustrations emerge when students rely on LLMs for lab implementations? (4) How do students transfer LLM-assisted coding skills to independent project work?

The rest of this paper is organized as follows. Section 2 reviews related work on AI-assisted programming education, pedagogical frameworks, and NLP teaching challenges. Section 3 describes our methodology, including the course context, Vibe Coding framework, and data collection methods. Section 4 presents quantitative and qualitative results from student feedback. Section 5 discusses implications for NLP education and broader CS pedagogy. Finally, Section 6 concludes with key lessons learned and recommendations for practitioners.

\section{Background and Related Work}

\subsection{AI-Assisted Programming in Education}

The emergence of AI coding assistants has sparked vigorous debate in computer science education. Early concerns focused on academic integrity and whether students would develop fundamental programming skills \cite{leinonen2023comparing, finnie2022robot}. However, recent work suggests that with appropriate scaffolding, AI tools can enhance learning outcomes \citet{denny2023conversing} \citet{macneil2023experiences}.

\citet{denny2023conversing} found that students using GitHub Copilot completed programming tasks faster with comparable quality, though struggled more with conceptual questions. \citet{kazemitabaar2023studying} observed that novice programmers using LLMs showed increased engagement but decreased learning on transfer tasks. These findings highlight the critical need for pedagogical frameworks that leverage AI assistance while preserving conceptual development.

\subsection{Pedagogical Approaches and Vibe Coding}

Several frameworks have emerged for integrating AI into programming education. \citet{sarsa2022automatic} propose treating AI assistants as ``pair programmers'' requiring critical evaluation. \citet{prather2023robots} advocate for explicit instruction on prompt engineering and output verification. \citet{macneil2023experiences} emphasize the importance of metacognitive scaffolding when students use generative AI.

Within this emerging landscape, we introduce \textbf{Vibe Coding} as a pedagogical approach that explicitly sanctions and structures LLM use in programming education. The term ``vibe coding'' informally refers to the practice of generating code through conversational interaction with LLMs, relying on natural language descriptions rather than manual syntax construction. While this practice has become widespread among students, it has typically been viewed by educators as either a form of academic misconduct or an unavoidable reality to be grudgingly accepted.

Our approach reframes vibe coding as a legitimate and valuable learning activity when properly scaffolded. We define Vibe Coding (capitalized to distinguish the pedagogical framework from informal practice) as a structured educational approach with three core components: (1) sanctioned and encouraged use of LLMs for code generation during learning activities, (2) mandatory documentation of the prompting process to make AI interaction visible and reflectable, and (3) assessment focused on conceptual understanding and critical evaluation rather than independent code production. This framework builds on existing work by explicitly separating implementation (delegated to LLMs) from conceptual understanding (assessed through reflection), while requiring students to maintain detailed logs of their AI interactions.

\subsection{NLP Education Challenges}

Teaching NLP presents unique challenges: students must understand linguistic theory, statistical methods, and software engineering simultaneously. Traditional approaches often overwhelm students with implementation complexity, obscuring conceptual learning \cite{pannitto2021teaching}.

Recent workshops on Teaching NLP have highlighted the need for pedagogical innovation as the field rapidly evolves \cite{teachingnlp2021proceedings}. The rise of large language models as both objects of study and tools for learning creates unprecedented opportunities to redesign NLP curricula.

\subsection{Assessment in AI-Augmented Learning}

Traditional code-based assessment becomes problematic when students can generate functional implementations with minimal understanding. Alternative assessment methods, including concept inventories, design artifacts, and reflective writing have shown promise for evaluating deeper learning \cite{cunningham2017teaching}.

Our reflection-based assessment approach draws on this literature, asking students to explain design decisions, analyze results, and critique LLM outputs rather than simply producing working code.

\section{Methodology}

\subsection{Course Context}

We implemented Vibe Coding in a senior-level undergraduate NLP course (Level 8) at the College of Computer and Information Sciences, King Saud University, Riyadh, Saudi Arabia, during the Fall 2025 semester. The course enrolled 19 students majoring in Information Technology (response rate: 100\%). Students had prerequisite knowledge of Python programming, data structures, and machine learning concepts. The course description states: "In this course, students will be exposed to methods for processing human language text and the underlying computational properties of natural languages. Students will explore natural language knowledge at different levels including phonetics, morphology, syntax, semantics, pragmatics and discourse levels. The course also introduces students to the evaluation techniques in the field of human language technologies. In addition to building applications to process written and/or spoken language.''\footnote{The course followed the structure and content of \textit{Speech and Language Processing} (3rd edition draft) by Jurafsky and Martin, focusing on Volume I: Large Language Models, available at \url{https://web.stanford.edu/~jurafsky/slp3/}}

The course comprised:
\begin{itemize}
    \item 12 weeks of lectures (2 hours each) covering NLP fundamentals
    \item 7 hands-on labs (2 hours each) following each lecture
    \item 1 multi-phase final project (Arabic Physical IQA dataset creation and model training)
\end{itemize}

\subsection{The Vibe Coding Framework}

\subsubsection{Lab Structure}

Each lab followed a consistent structure:

\begin{enumerate}
    \item \textbf{Lecture (2 hours)}: The instructor provided comprehensive coverage of the theoretical foundations and key NLP concepts for each topic.
    
    \item \textbf{Lab Practice (2 hours)}: Students worked through structured tasks using LLMs for code generation. They were explicitly permitted and encouraged to use ChatGPT, Claude, Gemini or other LLMs to implement the concepts covered in the lecture.
    
    \item \textbf{Prompt Log Documentation}: Students documented their prompts directly in the lab notebook before each task, recording the prompts they used, LLM responses received, and any modifications made to the generated code.
    
    \item \textbf{Critical Reflection}: Students completed a Google Form with questions probing conceptual understanding, design decisions, and analysis of results.
\end{enumerate}

Lab topics included:
\begin{enumerate}
    \item Introduction to NLP Preprocessing - Tokenization and Vocabulary Building
    \item POS Tagging and NER Analysis
    \item Text Classification
    \item N-Gram Language Models
    \item Word Embedding
    \item Fine-tuning Transformers
    \item In-Context Learning (Zero/Few-Shot)
\end{enumerate}

\subsubsection{Assessment Design}

Grading allocation shifted dramatically from traditional approaches:
\begin{itemize}
    \item \textbf{Code Output: 20\%} (functionality only; style irrelevant)
    \item \textbf{Prompt Log: 30\%} (quality of prompts, iteration process)
    \item \textbf{Critical Reflection: 50\%} (conceptual understanding, analysis depth)
\end{itemize}

Prompt logs were evaluated based on the clarity and specificity of prompts, evidence of iterative refinement, and completeness of documentation.
Reflection questions varied by lab but consistently targeted higher-order thinking. Examples from different labs included:
\begin{itemize}
    \item \textit{Lab 1 (Tokenization)}: ``What is the difference between the English and Arabic tokenization?'' ``What is a common issue with simple space-based tokenization, as seen in the expected output (e.g., 'dog.')?''
    \item \textit{Lab 2 (POS/NER)}: ``Which words in the text are tagged differently between the Penn Treebank tags and the Universal POS tags? List at least 2 examples and explain why the tagging schemes might differ in their classification.''
    \item \textit{Lab 3 (Text Classification)}: ``Compare the performance of all four ML models (Multinomial NB, Logistic Regression, Linear SVM, XGBoost) across different evaluation metrics. Which model performed best and why do you think this occurred?''
    \item \textit{Lab 4 (N-Grams)}: ``What is the difference in coherence between unigram, bigram, and trigram generated text? Provide examples of generated verses.''
    \item \textit{Lab 5 (Embeddings)}: ``Which model had the largest vocabulary and/or longest training time, and why?'' ``Suppose you switch min\_count from 5 to 2 for all models. What trade-offs do you expect in neighbor coherence and analogy accuracy?''
    \item \textit{Lab 6 (Fine-tuning Transformers)}: ``Which of the three architectures (Encoder-only, Encoder-decoder, or Decoder-only) would you recommend as a starting point for a new task like sarcasm detection, and why?''
    \item \textit{Lab 7 (In-Context Learning)}: Students explored zero-shot and few-shot prompting strategies with large language models.
\end{itemize}

\subsubsection{Final Project}

Students worked in small teams on a substantial four-phase project localizing the Physical Interaction Question Answering (PIQA) dataset to Arabic. The project spanned 12 weeks and involved:

\textbf{Phase 1 (Translation)}: Teams translated assigned PIQA items into Modern Standard Arabic while preserving minimal contrasts between solution pairs. Each item contained a goal and two candidate solutions (exactly one correct). Students documented translation challenges and cultural adaptations in detailed notes.

\textbf{Phase 2 (Corpus Analysis)}: Teams analyzed their Arabic corpus through length distributions, vocabulary and part-of-speech patterns, and identification of frequent contrasts (temporal, spatial, negation markers). They tagged these contrasts to define evaluation slices for later analysis.

\textbf{Phase 3 (Model Development)}: Teams implemented baselines (random, majority, lexical heuristics) and trained machine learning models including a linear classifier and transformer-based models (Arabic or multilingual). They conducted at least two ablation studies examining factors such as tokenization approaches or data augmentation strategies.

\textbf{Phase 4 (Evaluation)}: Teams performed comprehensive evaluation with overall accuracy metrics and slice-specific results for tagged contrasts. They conducted error analysis on 10-15 failed examples, categorized errors, proposed fixes, and validated at least one fix through experimentation.

Students were explicitly permitted to use LLMs for translation assistance, code generation, and debugging, but were required to document all AI interactions and conduct human quality assurance on all outputs. The project concluded with team presentations demonstrating their complete pipeline and findings.

\subsection{Data Collection}

At semester's end, we administered a comprehensive questionnaire covering:
\begin{itemize}
    \item General course experience (3 items)
    \item Vibe Coding process (4 items)
    \item Assessment structure (3 items)
    \item Project structure and LLM use (3 items)
    \item Open-ended feedback on strengths, challenges, and suggestions
\end{itemize}

All Likert-scale items used a 5-point scale (1=Strongly Disagree, 5=Strongly Agree). The questionnaire was anonymous to encourage honest feedback. We computed descriptive statistics (means, standard deviations) for quantitative items and conducted thematic analysis on open-ended responses to identify recurring patterns in student experiences.

\section{Results}

\subsection{Quantitative Findings}

\subsubsection{General Course Experience}

Students reported overwhelmingly positive course experiences (Table \ref{tab:general}). Course relevance received particularly strong endorsement (M=4.68, SD=0.67), suggesting students perceived the material as applicable to current industry trends. The balance between theory and practice was well-received (M=4.58, SD=0.69), as was confidence in applying core NLP concepts (M=4.00, SD=0.75).

\begin{table}[t]
\centering
\small
\begin{tabular}{lcc}
\toprule
\textbf{Item} & \textbf{Mean} & \textbf{SD} \\
\midrule
Course material relevant & 4.68 & 0.67 \\
Theory/practice balance & 4.58 & 0.69 \\
Confident in NLP concepts & 4.00 & 0.75 \\
\bottomrule
\end{tabular}
\caption{General course experience (N=19, 1-5 scale)}
\label{tab:general}
\end{table}

\subsubsection{Vibe Coding Process}

The Vibe Coding approach received strong support across all dimensions (Table \ref{tab:vibe}). Students found LLM use engaging (M=4.42, SD=0.84) and believed labs successfully taught conceptual understanding despite using LLMs for implementation (M=4.42, SD=0.69).

The mandatory prompt log was viewed favorably (M=3.79, SD=1.13), though with more variability, suggesting differential benefits across students. Notably, students felt the labs effectively taught critical evaluation of LLM outputs (M=4.37, SD=0.76).

\begin{table}[t]
\centering
\small
\begin{tabular}{lcc}
\toprule
\textbf{Item} & \textbf{Mean} & \textbf{SD} \\
\midrule
More engaging than traditional & 4.42 & 0.84 \\
Taught critical evaluation & 4.37 & 0.76 \\
Taught concepts despite LLMs & 4.42 & 0.69 \\
Prompt log useful & 3.79 & 1.13 \\
\bottomrule
\end{tabular}
\caption{Vibe Coding process ratings (N=19)}
\label{tab:vibe}
\end{table}

\subsubsection{Assessment Structure}

Students responded positively to the reflection-based assessment (Table \ref{tab:assessment}). The critical reflection questions were perceived as fair (M=4.21, SD=1.08). The grading weight shift was deemed appropriate (M=4.26, SD=0.99).

However, time allocation emerged as a concern: while the modal response was 5 (sufficient time), mean=3.53 (SD=1.43) indicates significant disagreement, with several students finding labs rushed.

\begin{table}[t]
\centering
\small
\begin{tabular}{lcc}
\toprule
\textbf{Item} & \textbf{Mean} & \textbf{SD} \\
\midrule
Reflection questions fair & 4.21 & 1.08 \\
Grading shift appropriate & 4.26 & 0.99 \\
Time allocation sufficient & 3.53 & 1.43 \\
\bottomrule
\end{tabular}
\caption{Assessment structure ratings (N=19)}
\label{tab:assessment}
\end{table}

\subsubsection{Project Experience}

Students viewed the final project as a valuable application opportunity (M=4.37, SD=0.83), though many found the scope challenging relative to available time (M=3.84, SD=1.26). Notably, 89\% of students used LLMs in their project work (Table \ref{tab:project}).

\begin{table}[t]
\centering
\small
\begin{tabular}{lcc}
\toprule
\textbf{Item} & \textbf{Mean} & \textbf{SD} \\
\midrule
Good skills application & 4.37 & 0.83 \\
Appropriate scope & 3.84 & 1.26 \\
Used LLMs in project & 4.53 & 1.17 \\
\bottomrule
\end{tabular}
\caption{Project experience ratings (N=19)}
\label{tab:project}
\end{table}

\subsection{Qualitative Findings}

\subsubsection{Reduced Cognitive Load Enables Deeper Learning}

The most prominent theme in student self-reports was that offloading implementation to LLMs freed mental resources for conceptual engagement. It is important to note that these findings reflect student perceptions rather than objective cognitive load measurements. Some students comments: \textit{``It made coding feel comfortable and with low pressure''} (S1); \textit{``Making me understand concepts rather than finding errors on my code''} (S2). Multiple students contrasted this favorably with traditional labs where debugging consumed disproportionate time. The ability to ``try different ideas and immediately see the results'' enabled iterative experimentation that deepened understanding.

\subsubsection{Prompt Engineering as Transferable Skill}

Students recognized that learning to effectively communicate with LLMs was itself valuable: \textit{``Getting better at writing prompts with time since we are instructed to practice it every lab''} (S4). The mandatory prompt log served as both accountability mechanism and learning scaffold.

\subsubsection{Verification and Trust Challenges}

A recurring frustration was verifying LLM outputs: \textit{``Sometimes the LLM would give answers that weren't completely accurate or would go off track, so I had to spend extra time fixing or verifying the results''} (S3). This highlights a pedagogical challenge: students need foundational knowledge to critically evaluate LLM outputs, creating a bootstrapping problem. Some struggled specifically with Arabic NLP models where LLM performance was less reliable.

\subsubsection{Time Pressure Persists}

Despite the pedagogical intent to reduce implementation time, many felt rushed: \textit{``Increase the lab time. The current duration is really short, and it's difficult to complete all the tasks properly''} (S3). Interestingly, LLM use didn't eliminate time pressure, it shifted it from debugging to understanding, verifying, and documenting AI-generated solutions.

\subsubsection{Strategic LLM Use in Projects}

Students demonstrated sophisticated understanding of when to leverage LLMs in final projects: debugging and error resolution, complex algorithm implementation, environment setup, and iterative guidance. Notably, students recognized LLMs as supplements: \textit{``I learned how to combine my own technical skills with AI tools in a strategic way---using LLMs as support, not as a replacement''} (S16).

\section{Discussion}

\subsection{Implications for NLP Education}

Our findings suggest that Vibe Coding can successfully shift student attention from syntactic details to conceptual understanding, based on student self-reports and qualitative feedback. However, these perceptions have not been validated through controlled cognitive load measurements, and it remains possible that reduced implementation effort does not directly translate to deeper conceptual processing. As noted in our future work directions, rigorous investigation using established cognitive load instruments is needed to substantiate these preliminary observations. High ratings for engagement and conceptual learning, combined with qualitative evidence of deep reflection, indicate students can learn NLP principles while using LLMs for implementation. However, several conditions appear necessary: (1) \textbf{Assessment alignment} with conceptual understanding rather than code quality, (2) \textbf{Prompt documentation} to encourage strategic LLM interaction, (3) \textbf{Verification support} through test cases, peer review, or instructor checkpoints, and (4) \textbf{Adequate time} for understanding and documenting AI-generated solutions.

\subsection{The Verification Challenge}

A central tension emerged: students need foundational knowledge to evaluate LLM outputs, yet acquiring that knowledge traditionally requires implementation practice. Possible resolutions include: (1) staged approaches beginning with traditional implementation for fundamentals, (2) hybrid labs mixing manual and LLM-assisted tasks, (3) explicit verification instruction, and (4) peer review where students compare and critique solutions.

\subsection{Limitations}

Several factors limit generalizability: small sample size (N=19) from a single institution, senior undergraduates with programming prerequisites, NLP's suitability for high-level API usage, challenges specific to Arabic text processing, and timing in Fall 2025 when LLM capabilities were mature. Additionally, we did not control for individual differences in prior LLM experience or programming proficiency, which may have influenced student outcomes and perceptions. Replication across diverse contexts is essential. Furthermore, we cannot rule out the possibility that students used LLMs to assist with their reflection responses, which may affect claims about demonstrated conceptual understanding. Future work could explore proctored or oral assessment methods to address this concern.
\subsection{Broader Implications}

Our findings contribute to AI-augmented learning literature: the shift from code artifacts to conceptual reflections parallels movements toward authentic assessment as AI tools become ubiquitous. Programming competency may need redefinition emphasizing problem decomposition, algorithm selection, output verification, and strategic AI collaboration rather than raw coding ability. By explicitly teaching thoughtful AI tool use rather than prohibiting them, we better prepare students for ethical professional practice.

\section{Conclusion and Lessons Learned}

This paper presents the first systematic evaluation of ``Vibe Coding'', a pedagogical approach embracing LLM assistance in NLP education while maintaining rigorous conceptual assessment. Our findings demonstrate that when properly structured, LLM-assisted learning can redirect student attention from low-level implementation to high-level conceptual understanding.

\subsection{Key Lessons}

\textbf{Lesson 1: Assessment drives learning}. The shift in grading weight from code to reflection was critical. Students engaged deeply with concepts because that is what was assessed.

\textbf{Lesson 2: Prompt logging encourages metacognition}. The mandatory prompt log served important functions, though future iterations might improve through exemplar prompts or peer review.

\textbf{Lesson 3: Time pressure persists}. LLMs transform rather than eliminate time constraints. Students need time to understand, verify, and reflect on AI-generated solutions.

\textbf{Lesson 4: Verification is non-trivial}. Teaching students to critically evaluate LLM outputs requires explicit instruction and support, the most significant remaining challenge.

\textbf{Lesson 5: Transfer happens}. Students successfully applied Vibe Coding strategies to independent project work, demonstrating generalizable skills.

\subsection{Future Work}

Several research directions emerge from this initial exploration of Vibe Coding:

\textbf{Longitudinal studies}: How does Vibe Coding affect long-term learning outcomes and professional performance? Tracking students across multiple semesters and into industry positions would reveal whether the conceptual foundations developed through Vibe Coding translate to sustained competence in advanced courses and real-world contexts.

\textbf{Controlled experiments}: Rigorous comparative studies between Vibe Coding and traditional approaches are needed to establish causal evidence of effectiveness. Such studies should measure conceptual understanding through validated instruments, implementation skills with and without AI assistance, and transfer learning to novel problems.

\textbf{Cognitive load measurement}: More rigorous investigation of how LLM assistance affects cognitive load distribution during learning could reveal whether reduced syntax-related load genuinely frees cognitive capacity for conceptual processing, or whether students simply engage less deeply overall.

\textbf{Verification pedagogy}: The challenge of teaching students to critically evaluate LLM outputs emerged as a central concern. Future work should develop and test specific instructional approaches such as graduated scaffolding with explicit verification strategies, comparative analysis of multiple LLM-generated solutions, and structured peer review protocols.

\textbf{Diverse populations}: Testing Vibe Coding with different student populations (introductory vs. advanced, varying programming backgrounds, different cultural contexts) would establish the generalizability of the approach and identify necessary adaptations for different learning contexts.

\textbf{Cross-domain application}: Exploring whether Vibe Coding principles transfer effectively to other computing domains beyond NLP (e.g., algorithms, systems programming, databases) would reveal the broader applicability of the framework and help identify domain-specific challenges in balancing conceptual understanding with practical implementation.

\subsection{Concluding Thoughts}

Large Language Models are reshaping software development and computing education. Rather than resisting this transformation, Vibe Coding embraces it as an opportunity to refocus education on enduring conceptual knowledge and higher-order thinking skills. Our findings suggest this approach is viable and could represent a productive path forward for NLP education in the age of AI.

The goal is not to eliminate programming skill but to redefine it for an AI-augmented world where critical thinking about problems, choosing appropriate approaches, and verifying solutions matters more than raw syntax fluency. As one student expressed: ``I learned how to combine my own technical skills with AI tools in a strategic way---using LLMs as support, not as a replacement.'' This represents exactly the kind of sophisticated AI literacy we should cultivate in the next generation of NLP practitioners.

\bibliography{references}

\end{document}